\documentclass[10pt, a4paper]{article}
\usepackage{lrec2022} 
\usepackage{multibib}
\newcites{languageresource}{Language Resources}
\usepackage{graphicx}
\usepackage{tabularx}
\usepackage{soul}
\usepackage{lingmacros}
\usepackage[justification=centering]{caption}
\usepackage{times}
\usepackage{latexsym}
\usepackage{graphicx}
\usepackage{booktabs,multirow}
\usepackage{amssymb}
\usepackage{tabularx}
\usepackage{bm}
\usepackage{amsmath,mathtools}
\usepackage{amsfonts}
\usepackage{longtable}
\usepackage{wrapfig}
\usepackage{enumitem}
\usepackage{pifont}
\usepackage{longtable}
\usepackage{mwe}
\usepackage{makecell}
\usepackage{dblfloatfix}

\usepackage{varwidth}
\usepackage{array}

\usepackage{newfloat}
\usepackage{caption}
\usepackage{multicol}

\DeclareFloatingEnvironment[placement={!ht},name=List]{mylist}

\usepackage[switch]{lineno}
\usepackage{lipsum}
\usepackage{titlesec}
\titleformat{\section}{\normalfont\large\bfseries\center}{\thesection.}{1em}{}
\titleformat{\subsection}{\normalfont\SmallTitleFont\bfseries\raggedright}{\thesubsection.}{1em}{}
\titleformat{\subsubsection}{\normalfont\normalsize\bfseries\raggedright}{\thesubsubsection.}{1em}{}
\renewcommand\thesection{\arabic{section}}
\renewcommand\thesubsection{\thesection.\arabic{subsection}}
\renewcommand\thesubsubsection{\thesubsection.\arabic{subsubsection}}

\usepackage{epstopdf}
\usepackage[utf8]{inputenc}

\usepackage{hyperref}
\usepackage{xstring}

\usepackage{color}

\usepackage{calc}
\title{\normalfont{\textbf{LPAttack: A Feasible Annotation Scheme for Capturing \\ Logic Pattern of Attacks in Arguments}}}

\name{Farjana Sultana Mim$^{1}$, Naoya Inoue$^{2,3}$, Shoichi Naito$^{1,3,4}$, Keshav Singh$^{1}$, Kentaro Inui$^{1,3}$} 

\address{
$^{1}$Tohoku University, 
$^{2}$Stony Brook University
$^{3}$RIKEN, 
$^{4}$Ricoh Company, Ltd.\\
         naoya.inoue.lab@gmail.com, inui@tohoku.ac.jp\\
         \{mim.farjana.sultana.t3, naito.shoichi.t1, singh.keshav.t4\}@dc.tohoku.ac.jp\\}

\abstract{
In argumentative discourse, persuasion is often achieved by refuting or attacking others arguments. Attacking is not always straightforward and often comprise complex rhetorical moves such that arguers might agree with a logic of an argument while attacking another logic.
Moreover, arguer might neither deny nor agree with any logics of an argument, instead ignore them and attack the main stance of the argument by providing new logics and presupposing that the new logics have more value or importance than the logics present in the attacked argument.
However, no existing studies in the computational argumentation capture such complex rhetorical moves in attacks or the presuppositions or value judgements in them. 
In order to address this gap, we introduce LPAttack, a novel annotation scheme that captures the common modes and complex rhetorical moves in attacks along with the implicit presuppositions and value judgements in them. 
Our annotation study shows moderate inter-annotator agreement, indicating that human annotation for the proposed scheme is feasible.
We publicly release our annotated corpus and the annotation guidelines.
 \\ 
 \newline  }

\begin{document}


\maketitleabstract

\section{Introduction}

Argumentation plays a central role in human communication where refuting or attacking other's arguments is a common strategy of persuasion~\cite{walton2010persuation}.
\emph{Attacks} in arguments can have different modes~\cite{walton2009objections,cramer2018directionality} and often comprise complex rhetorical moves, e.g., one might attack the \emph{conclusion} (i.e., statement expressing position or belief of the arguer) while agreeing with a \emph{premise} (i.e., statement providing support or reason for the conclusion) of the argument~\cite{afantenos2014counter}.

\begin{figure*}[!t]
\begin{center}
\includegraphics[width=14.5cm, height=6.9cm]{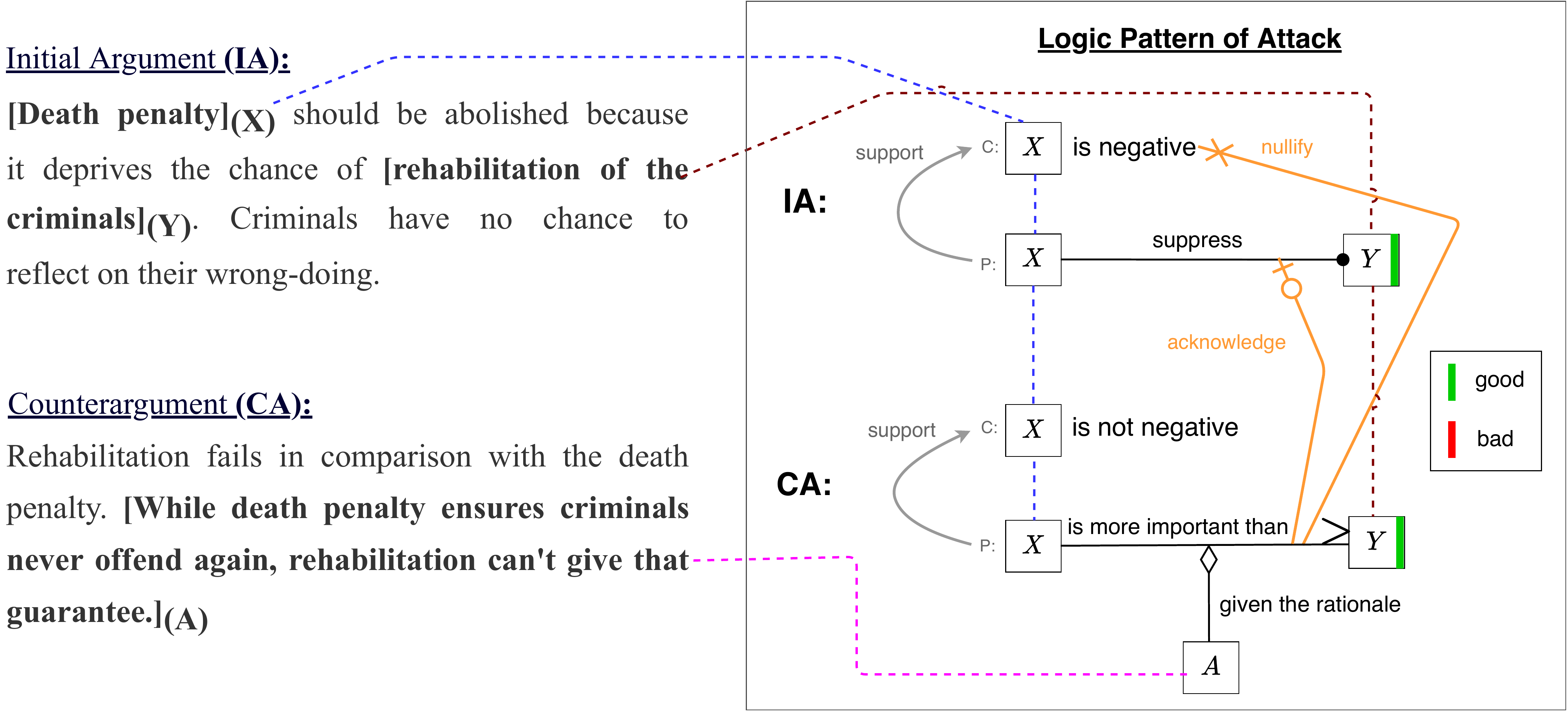}
\caption{An example of logic pattern of attack of a debate captured by \\ the proposed LPAttack annotation scheme.}
\label{Fig:lpattack-example}
\end{center}
\end{figure*}

Consider an example debate in Fig.~\ref{Fig:lpattack-example}, where two argumentative speeches are given by each opposing team of the debate.
%
%
In the debate, the counterargument (CA) does not deny the \emph{premise} of the initial argument (IA), i.e., \emph{death penalty deprives the chance of rehabilitation of the criminals}, instead she implicitly agrees with it while she denies the \emph{conclusion} of the IA i.e., \emph{death penalty should be abolished} by giving more importance or value to the \emph{death penalty} than the \emph{rehabilitation of the criminals}.
Although this value judgement is implicit in the CA speech, CA explicitly provides a reason behind her value judgement (bold text in CA).
Automatically identifying such internal logic patterns can help a wide range of natural language processing (NLP) applications.
For example, in an educational domain, this can help machines diagnose learners' arguments and provide feedback to the learners.

Prior studies in NLP that focused on \emph{attacks} in arguments mainly worked on classification of argumentative relations (e.g., support, attack, neural), identifying attackable points in arguments or counterargument generation \cite{stab2014annotating,deguchi2019argument,kobbe2019exploiting,jo2021classifying,walton2008argumentation,jo2020detecting,wachsmuth2018retrieval,hua2019argument,reisert2019riposte,alshomary2021argument,jo2021knowledge}.
Comparatively, less attention has been paid to identifying the logic pattern of attacks in arguments.

Although some recent studies \cite{reisert2018feasible,jo2021classifying} developed annotation schemes and logical mechanisms to capture the reasoning process behind support and attack relations where they exploited implicit causal links and sentiments, these studies didn't capture other implicit information e.g., presupposition or value judgements in arguments that also contribute to the underlying logical structure of attacks.
Furthermore, none of these studies capture the modes of \emph{attacks} (e.g, whether the counterargument denies the conclusion or the premise of the attacked argument) and the complex rhetorical moves in them (e.g., agreeing with a premise while attacking the conclusion).


In order to address these gaps, we introduce LPAttack (\textbf{L}ogic \textbf{P}atterns of \textbf{Attack}), a new annotation scheme that captures common modes of attacks and complex rhetorical moves in them as well as the implicit information and value judgements that contribute to the logical structure of attacks.
Fig.~\ref{Fig:lpattack-example} shows an example annotation.
The logic pattern of IA speech is represented by our logic pattern, which can be interpreted as follows: \emph{\ul{death penalty}~(=X) is considered a negative thing because \ul{death penalty} suppresses \ul{chance of rehabilitation of the criminals}~(=Y), something good}.
The logic pattern of CA speech then represents their value judgement on \emph{death penalty} and \emph{chance of rehabilitation of the criminals}, where more value is given to \emph{death penalty}.
This value judgement then attacks the conclusion of IA.
Given that information, one can understand how and which part of the IA is attacked by CA.

Our contributions can be summarized as follows:
%
%
%
\begin{itemize}
    \item We introduce LPAttack, a novel annotation scheme that captures the common modes and complex rhetorical moves in attacks along with the implicit information, presuppositions or value judgements in them~(\S\ref{sec:annotation_scheme}).
    \item We conduct an annotation study using the proposed scheme that yields moderate agreement between two annotators indicating the feasibility of the human annotation for the scheme~(\S\ref{sec:annotation_study}).
    \item We provide the annotated corpus consisting of logic patterns of attacks of 250 debates and the annotation guidelines as a publicly available resource to encourage future research\footnote{Our annotated corpus and annotation guidelines are publicly available at http://.}.
\end{itemize}

\section{Related Work}
Computational analysis of argumentation has gained considerable attention in recent years due to its importance in many NLP application such as essay scoring, argumentative writing support system, providing educational feedback etc. Common lines of work in this area include argumentative units (e.g., claim, premise) identification \cite{levy2014context,rinott2015show,stab2014annotating}, argumentative relations (e.g., support, attack, neural) classification \cite{peldszus2015annotated,cocarascu2017identifying,niculae2017argument,stab2014annotating,deguchi2019argument,kobbe2019exploiting,jo2021classifying}, qualitative assessment of arguments \cite{persing2010modeling,persing2013modeling,persing2014modeling,persing2015modeling,persing2016modeling,rahimi2015incorporating,wachsmuth2016using,habernal2016argument,wachsmuth2017computational,mim2019unsupervised,mim2021corruption} and retrieval or generation of counterargument \cite{hua2018neural,wachsmuth2018retrieval,hua2019argument,reisert2019riposte,alshomary2021argument,jo2021knowledge}

Lately, researchers have started focusing on one of the complex and challenging facet of argument analysis, i.e., capturing or explicating the encapsulated knowledge in arguments (e.g., causal knowledge, commonsense knowledge, factual knowledge) which are often implicit \cite{habernal2017argument,hulpus2019towards,becker2019implicit,becker2020explaining,al2020end,becker2021reconstructing,becker2021co,singh2021exploring,saha2021explagraphs}.
Although for deeper understanding of argumentation, we also need to comprehend the underlying reasoning patterns of arguments, less attention has been paid to representing such underlying reasoning patterns and explicating the implicit information that contribute to these patterns.

We focus on this gap and address the problem of explicating internal logic pattern of attacks in arguments that comprise complex rhetorical moves and implicit causal information, sentiments, presuppositions or value judgements.
Our inspiration for designing such annotation scheme comes from Walton's argumentation schemes \cite{walton2008argumentation} which represent the common reasoning structures in arguments.
To give an example, Walton's scheme of \emph{Argument from Negative Consequences} has the conclusion \emph{A should not be brought about} which is supported by the premise \emph{if A is brought about, then bad consequences will occur}.
Although Walton's schemes explicate the unstated assumptions or propositions as a form of reasoning pattern, they are not intended for capturing the logic pattern of \emph{attacks}, i.e., how an argument is attacked by a counterargument. 
It should be noted that each of the Walton's scheme has a set of critical questions (CQs) associated with it that are used to judge if an argument fitting a scheme is good or fallacious. 
Some CQs for the above scheme are \emph{How strong is the likelihood that the cited consequences will occur?},  \emph{Are there other opposite consequences that should be taken into account?}. 
However, the critical questions in Walton's schemes only specify the attackable points in an argument, they do not represent the reasoning pattern of attacks.

Some recent studies adopted Walton's schemes to represent the logic behind support and attack relations.
One of these studies \cite{reisert2018feasible} developed an annotation scheme that uses argument templates to capture reasoning patterns behind support and attack relations. 
Another study \cite{jo2021classifying} composed a set of rules specifying logical mechanisms that signal the support or attack relation. 
While these studies identified implicit causal reasoning, sentiments or factual contradiction in attacks, they did not capture other implicit information e.g., contradictory causal reasoning, assumptions or value judgements that significantly contribute to the logical structure of attacks.   

One recent study \cite{saha2021explagraphs} created commonsense explanation graphs that explicates the commonsense reasoning process involved in inferring support and attack relations.
However, the focus of this study is commonsense explanation, not the reasoning pattern of \emph{attack} (or support). 
Therefore, although this study exploited implicit causal knowledge, the fine-grained implicit knowledge explicated in this study is not effective in representing the logical structure of \emph{attacks} which requires distinct coarse-grained implicit information e.g., contradictions or value judgements in arguments.

Besides, there is still no work in computational argumentation that captures the modes of attacks in arguments (e.g, whether the counterargument denies the conclusion or the premise of the attacked argument) or the complex rhetorical moves in them (e.g., agreeing with a premise while attacking the conclusion, providing a contradictory premise that leads to denying the conclusion etc.). Our work addresses these gaps by introducing an annotation scheme that can capture common modes of attacks, complex rhetorical moves in them as well as implicit causal reasoning, sentiments, presuppositions or value judgements that contribute to the logic pattern of attacks.

\begin{figure*}[!t]
\begin{center}
\includegraphics[width=14.5cm, height=5.8cm]{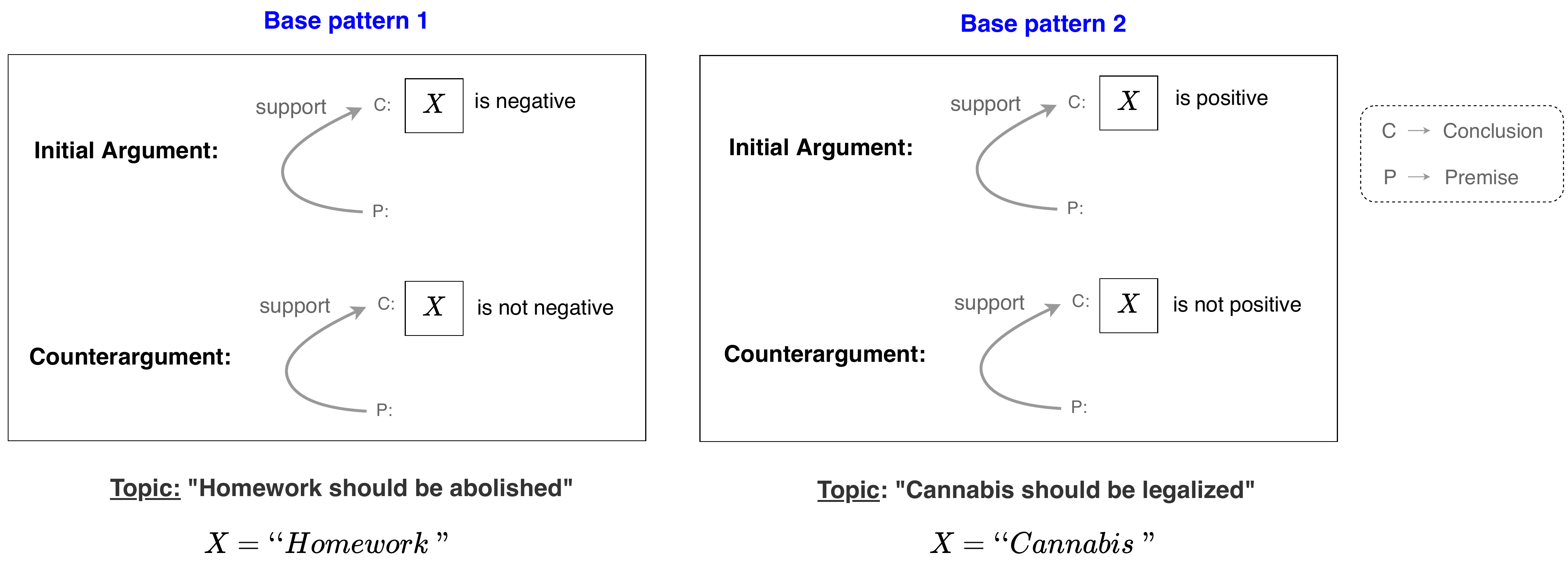} 
\caption{Base logic patterns with examples.}
\label{Fig:base-pattern}
\end{center}
\end{figure*}

\section{LPAttack Annotation Scheme}
\label{sec:annotation_scheme}
In our study, we hypothesize that the logic pattern of attacks in arguments are not uniformly distributed rather highly skewed and following this hypothesis, we develop our annotation scheme to capture the common logic pattern of attacks in arguments.

\subsection{Pre-Study and Scheme Design}
In order to examine what sort of strategic moves, assumptions or value judgements are common during attack, we first conduct a preliminary study where we perform qualitative analysis of how one argument attack another (see Appendix for further details).
For our pre-study, we select 35 debates from the TYPIC dataset\footnote{This dataset is publicly available} ~\cite{naito2022} which consists of multiple, diverse debate themes. Each debate comprise an argument and a counterargument conveyed by two opposing team of the debates.


This analysis of the internal structure of attacks provide insights into how we can represent the attacking logic so that human annotation is plausible.
Then based on these insights, we design our annotation scheme, define the annotation guidelines and formulate the task of capturing logic pattern of attacks in arguments.
In the following subsections, we describe our annotation scheme along with the findings in pre-study.

\subsubsection{Base Logic Patterns}
Generally, when people argue \emph{for} some belief, they show positive sentiment towards a concept of that belief. Conversely, when they argue \emph{against} some belief, they show negative sentiment towards a concept. For example, for the beliefs \emph{`homework should be abolished'} and \emph{`death penalty should be abolished'}, the arguers have \emph{against} stance and in both cases they have \emph{negative} sentiment towards the concepts \emph{homework} and \emph{death penalty}.
Now, counterarguments generally have the opposite stance and sentiment of initial argument e.g., the counterargument \emph{homework should not be abolished} have \emph{for} stance as well as \emph{non-negative or positive} sentiment towards the concept \emph{homework}.
No matter how diverse the topics of the arguments are, the sentiment towards a certain concept is generally dependent on these \emph{for} or \emph{against} stance.

Argumentation Schemes~\cite{walton2008argumentation}  highlighted this fact and extensively utilized the positive or negative sentiment of the arguer towards a certain concept or consequence.
Motivated by that, we design two base patterns (shown in Fig.~\ref{Fig:base-pattern}) for our scheme where the sentiments towards the main concept in the argument acts as the \emph{conclusion} of the argument. \emph{Base pattern 1} represents \emph{against} stance of the initial argument and therefore presents the logic: \{\emph{Initial Argument}: $X$ is negative; \emph{Counterargument}: $X$ is not negative\} where $X$ is a slot for the concept. 
Conversely, \emph{Base pattern 2} represents the case where initial argument has \emph{for} stance.
Both of the base patterns have two slots for premises, one in the initial argument and one in the counterargument.

\subsubsection{Relations and Attributes}
In order to capture the logic of \emph{premise} that will support the conclusion (i.e., the sentiment towards the central concept), we design a set of relations and attributes.
See Table~\ref{tab:template_set} in Appedix for the overview.

\paragraph{Causal relation}
Previous studies that worked on the representation of implicit reasoning behind support or attack mostly adopted Argumentation Schemes~\cite{walton2008argumentation} and have shown that a majority of the arguments can be represented by the implicit causal links \cite{reisert2018feasible,al2020end,jo2021classifying,singh2021exploring}. 
In our pre-study, we see the similar phenomena i.e., most of the logics in arguments that are attacked or acknowledged by counterarguments can be represented by causality.
For example, in Fig.~\ref{Fig:lpattack-example}, the logic of IA \emph{death penalty deprives the chance  of rehabilitation  of  the  criminals} can be represented with the ``suppress" causality i.e., \emph{\{death penalty, suppress, the chance  of rehabilitation of the criminals\}}.
We thus design our annotation scheme around two causal relations \emph{``promote"} and \emph{``suppress"} (henceforth, \emph{``base relations"}).

\paragraph{Value judgement}
We observe that one of the common reasoning during attack is based on value-judgements i.e., comparing two factors by giving more value or importance to one than the other. 
We have found two phenomena: (i) counterarguments give more importance to a certain concept of a logic while implicitly acknowledging the logic, as shown in Fig.~\ref{Fig:lpattack-example}; and (ii) counterarguments neither acknowledge nor deny any logic of the initial argument, instead ignore it and deny the conclusion of the initial argument by providing new reasons presupposing that the new reasons have more value. Consider the following example:

{\enumsentence{
\textbf{Initial Argument (IA)} \\
\emph{...homework should be abolished} 
\ul{(Conclusion)}\\
\emph{...if homework were to be abolished, we could have more free time. As a result, we could do what we really wanted like club activities...} \ul{(Premise)}\\
\textbf{Counterargument (CA)}\\
\emph{...if homework is abolished, a number of people who don’t study at all will increase.....To decrease a number of people who repeat years, homework is necessary...} \label{ex:2}}}

In Example~(\ref{ex:2}), CA neither affirms nor denies IA's logic, instead ignores it and provide new reasons that deny the conclusion ``homework should be abolished". CA presupposes that the value or importance of \emph{\{if homework is abolished, a number of people who don’t study at all will increase\}} is greater than the value of \emph{\{if homework were to be abolished, we could have more free time\}} and this presupposition is implicit in CA's argument.
In order to represent these two phenomena of value judgements, we create the relation \emph{``is more important or severe or has greater weight"}.

\paragraph{Contradiction}
We see another common attacking strategy in counterarguments i.e., providing contradictory logic that ultimately leads to denying the conclusion of the initial argument. One example of such scenario is given below:

{\enumsentence{
\textbf{Initial Argument (IA)} \\
\emph{..death penalty should be abolished} \ul{(Conclusion)}\\
\emph{...death penalty is causing brutalization of modern society......it validates the notion that the taking of someone’s life is a valid choice.....} \ul{(Premise)}\\
\textbf{Counterargument (CA)}\\
\emph{...death penalty sends a message that taking an innocent life will not be tolerated by a civilized society. Thus, it serves as an antidote to brutality...}} \label{ex:3}}

In the above example, while IA says: \emph{\{death penalty is causing brutalization of modern society\}}, CA says the opposite: \emph{\{it serves as an antidote to brutality\}}. CA's logic contradics IA's which leads to denying the conclusion of IA. In order to capture such contradictory logic, we invent the \emph{``contradiction"} relation.


\paragraph{Logic denial/agreement}
In order to explicitly represent the denial of a premise logic or conclusion, we create two relations \emph{``nullify"} and \emph{``limit"}. These relations are considered as the \emph{``attacking relations"} in our scheme. 
Besides, we represent agreeing with a logic by the relation ``acknowledgement".

\paragraph{Negation}
We observe that it is pretty common for counterarguments to \emph{negate} (explicitly or implicitly) certain logic, especially causal reasoning, by providing some rationales or conditions. 
Consider the following example:
{\enumsentence{
\textbf{Initial Argument (IA)} \\
\emph{...homework should be abolished} 
\ul{(Conclusion)}\\
\emph{...if students are always given homework, they will always be waiting for instructions.....homework should be abolished so that students can study on their own initiative....} \ul{(Premise)}\\
\textbf{Counterargument (CA)}\\
\emph{...students will hardly be able to study on their own without homework because continuous instructions or guidelines are needed for children to study or learn something new...} \label{ex:4}}}

In Example~(\ref{ex:4}), CA negates IA's logic \emph{\{homework should be abolished so that students can study on their own initiative\}} by saying \emph{\{students will hardly be able to study on their own without homework\}} and provides a reason behind it i.e., \emph{``continuous instructions or guidelines are needed for children to study or learn something new"}.
In order to represent such negation attribution and reasoning behind a logic we develop ``negation" attribute and ``rationale/condition" relation.

\paragraph{Mitigation}
Instead of completely negating a logic, counterarguments often express that the severity of such phenomena can be mitigated. Consider the following example: 

{\enumsentence{
\textbf{Initial Argument (IA)} \\
\emph{...death penalty should be abolished} 
\ul{(Conclusion)}\\
\emph{...death penalty causes executioner's suffering.....they feel that they are responsible themselves for killing the suspect....} \ul{(Premise)}\\
\textbf{Counterargument (CA)}\\
\emph{..executioner's stress can be reduced by making sure that would-be executioners are fully prepared for the job and have a good mental support system}}}

In this example, CA doesn't completely negate IA's logic \emph{\{death penalty causes executioner's suffering\}} instead partially negate it by saying \emph{\{executioner's stress can be reduced\}}.
We create ``mitigation" attribute to represent such partial negation attribution.

\subsubsection{Slot-filling}
Towards computationalizing the task of capturing logic pattern of attacks, we analyzed if it is possible to represent the logic patterns using only the information present in the argumentative texts (i.e., by not having any external commonsense concepts).
Our analysis suggested that it is fairly possible to represent the logic behind attack without writing any external commonsense concepts. We thus decided to formulate the task of slot filling as a text-span selection task: annotators choose slot fillers in the base patterns from the given arguments.

\subsection{Task Setting}
\label{task-setting}
The task of representing logic pattern of attack for a given argument and counterargument consists of the following steps:
\begin{enumerate}
    \item \emph{Selection of base logic pattern and slot-filling:} At first, a base logic pattern is selected based on the central stance of initial argument and then the slots of the pattern is filled with the central concept. 
    \item \emph{Selection of relations and attributes along with text-spans:} In this step, relations and attributes are chosen along with the text-spans from the given arguments to complete the base pattern by representing the logic of the premises.
\end{enumerate}
Since there is no fixed template for representing the logic of the premises in initial argument and counterargument of base pattern, one important question is how many relations, attributes and text spans should be chosen for premise representations and how to choose it. 
In this regard, we take a summarization approach i.e., we create one or two line mental summary of the counterargument (CA) considering its main points and find the logic in initial argument (IA) that is attacked by the CA. If CA attacks IA's conclusion instead of attacking any logic behind the conclusion, then we also create a one line mental summary of IA's main points.
Then, we choose suitable text spans, relations and attributes to represent that one or two line mental summary of CA, the attacked logic or mental summary of IA and how CA attacks IA (i.e., which part of the IA logic is denied and if CA agrees with any of the IA logic). We refer to these representations as \emph{CA-pattern}, \emph{IA-pattern} and \emph{attack-pattern}.

We set some constraints on how many relations and attributes can be selected for each of these three representations. For IA-pattern, at most two and for CA-pattern, at most three relations or attributes can be selected where using a base causal relation is mandatory for IA-pattern and using \emph{good} or \emph{bad} attributes are prioritized in both cases.
For representing the attack-pattern, choosing at least one attacking relation (i.e., nullify or limit) is mandatory. It should be noted that attack-pattern basically represents the relation between CA-pattern and IA-pattern or CA-pattern and the conclusion of IA.



We also set a constraint on how long the text spans should be. We specify that although text spans can be long up to two small sentences or one compound sentence, we should try to choose smaller text span (e.g., short phrase) as much as possible. We also specify that we should choose such text spans that when we read the patterns as a standalone logic (i.e., without reading the debates), they are understandable.


\section{Annotation Study}
\label{sec:annotation_study}

The key requirements for identifying logic pattern of attacks in arguments are two-fold: (i) identify as much logic pattern of attacks as possible and (ii) make human annotation feasible.
In order to verify whether our LPAttack scheme satisfies these requirements, we observe two metrics during annotation study: (i) coverage of the scheme and (ii) inter-annotator agreement (IAA).

\subsection{Setup}

Two expert annotators participate in the annotation study and annotate logic pattern of attacks independently using our annotation scheme\footnote{We use diagrams.net for annotation.}. 

We trained the annotators in a pilot annotation phase where both of the annotators were asked to annotate 20 debates. 
After the pilot annotation, we discussed the disagreements and, if needed, adjourned the annotation guidelines.
One issue that we observed in the pilot annotations is that when we read the annotated patterns as a standalone logic (i.e., without reading the debates), some of them didn't make complete sense because the chosen text spans had information gap. In order to reinforce that the logic patterns are understandable independently, we ask the annotators to write the text form of the logic patterns during our main annotation i.e., the way they read the logic patterns. For example, the text form of the logic pattern in Fig.~\ref{Fig:lpattack-example} is:
\begin{itemize} [label={}]
    \item IA: \emph{\{``death penalty" is negative\}} because \emph{\{``death penalty" suppress (``chance of rehabilitation of the criminals" which is good)\}}
    \item CA: \emph{\{``death penalty" is not negative\}} because \emph{\{``death penalty" is more important/severe/has greater weight than ``chance of rehabilitation of the criminals which is good" given the rationale/condition that  ``while executing prisoners is completely effective in ensuring...."\}}
\end{itemize}
We expect that writing the text form would serve as a double checking for the logic patterns and when the annotators would read it separately from the debates, they will understand if there is an information gap or if the logic pattern is self-sufficient. 


In our main annotation study, 50 debates are annotated by two annotators and 145 debates are annotated by a single annotator.
For coverage and IAA, we report the results of dual annotations on 50 debates.


\subsection{Coverage}
We ask the annotators to mark an attacking strategy as ``Not Applicable (NA)" if it cannot be represented by our scheme. We obtain 90\% (45/50) coverage for the LPAttack scheme. This result validates our hypothesis that logic pattern of attacks in arguments are not uniformly distributed rather highly skewed i.e., logic pattern of a wide range of attacks can be captured with a limited set of relations and attributes.

\subsection{Inter-annotator agreement (IAA)}
We measure the IAA for relations and attributes\footnote{We ignore calculating the agreement for the selection of base pattern since all the debates used in the annotation study has the same base pattern (i.e., base pattern 1).} using Cohen’s ($\kappa$) \cite{cohen1960coefficient}.
For the calculation of IAA, we consider IA-pattern, CA-pattern and attack-pattern (described in section \ref{task-setting}) as our markables.
Since we want to know how much the annotators agree on each of these logics as well as on the overall debate, we employ two strategies for the calculation of IAA: (i) calculate IAA considering each markables, (ii) concatenate these three markables to have a single representation of the whole debate and calculate IAA.

We obtain Cohen's $\kappa$ of 0.63 in case of (i) which indicates a substantial agreement and in case of (ii), we obtain 0.49 which indicates a moderate agreement.

Additionally, we evaluate whether the text spans are the same in cases where relations and attributes are agreed.
It should be noted that among the three markables, only IA-pattern and CA-pattern have text spans and therefore we consider these two markables for this matching calculation but follow the same strategy as above (i.e., (i) and (ii)).
In each of the markables, if all of the text spans exactly match, we call it \emph{exact-match}, if all of the text spans leniently (semi-exactly) match or some them have lenient matching while others have exact matching, we call it \emph{lenient-match}. 
We see that in case of (i), 68\% (47/69) text spans are similar \emph{(43\% (30/69) exact-match, 25\% (17/69) lenient-match)}.
For (ii), we obtain 46\% (12/26) match \emph{(19\% (5/26) exact-match, 6\% (7/26) lenient-match)}.

These results suggest that annotators can choose appropriate relations and attributes as well as text spans with a certain degree of reliability.

\begin{figure*}[!t]
\begin{center}
\includegraphics[width=14cm, height=9.5cm]{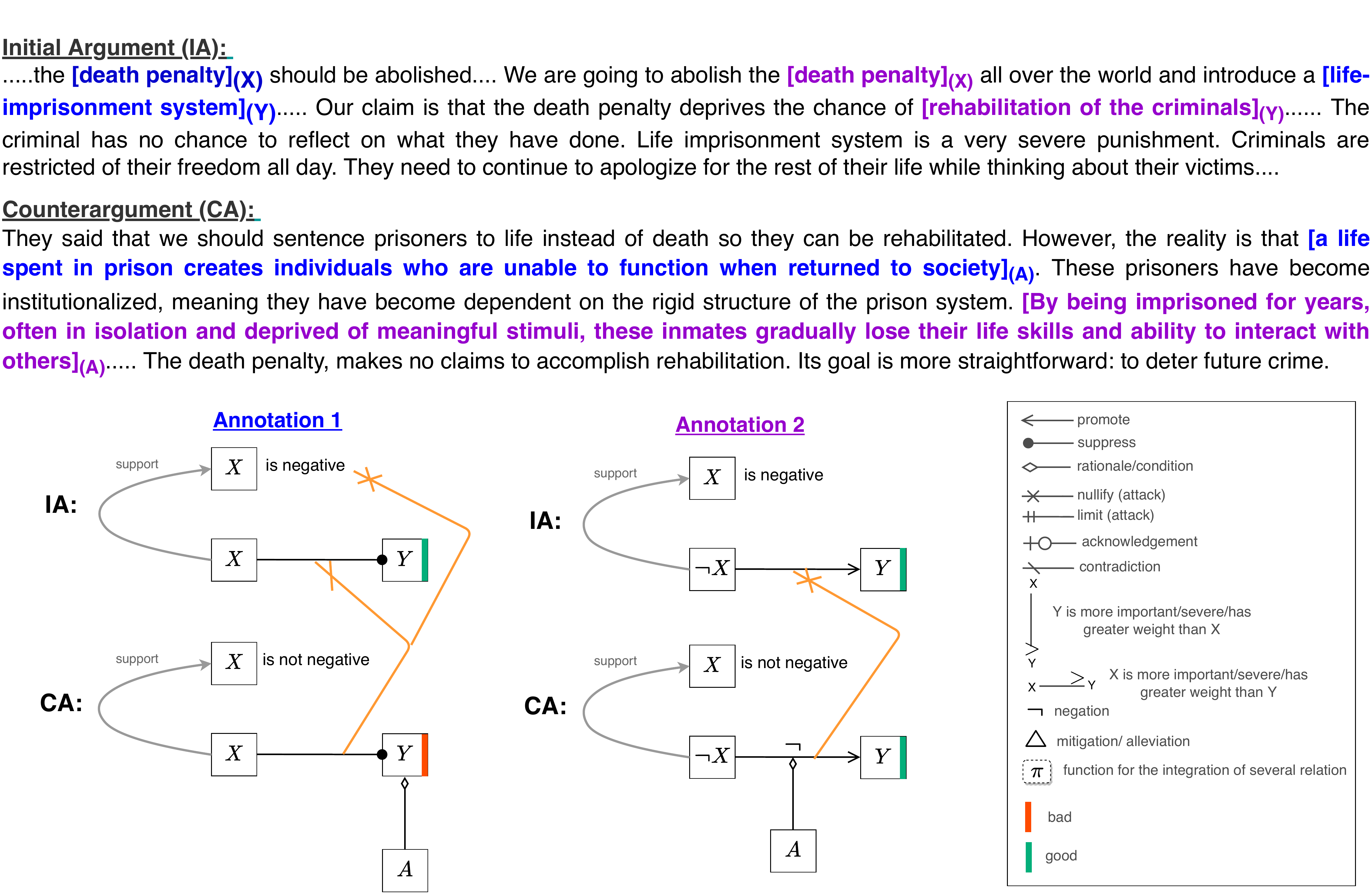} 
\caption{Example of debate where two annotators have different interpretation.}
\label{Fig:dif-interpret}
\end{center}
\end{figure*}

\subsection{Analysis of Annotations}
We perform a manual analysis of the annotations in order to examine the correctness of the logic patterns, disagreements between the annotators and common attacking strategies captured by these annotations.


\paragraph{Correctness of the logic patterns} We check how many annotated logic patterns are correct i.e., the logic patterns capture essence of the attacks and are understandable enough when read independently without reading the debates. We adopt the following strategy in this regard:
\begin{itemize}
    \item Exact logic pattern match between annotators \ding{254} mark it as a correct logic pattern
    \item Non-exact match between annotators  \ding{254} manually check the logic patterns and discuss with the annotators \ding{254} mark it as correct or incorrect based on the decision in the discussion
\end{itemize}
Following the above strategy, we find that 90\% (45/50) annotations of one annotator is correct where 86\% (43/50) annotations of the another annotator is correct.
Both the annotators had incorrect patterns for two of the debates. For four of the debates, one annotator chose ``NA" while other had correct patterns and for one of the debates one annotator chose ``NA" while other had incorrect pattern.
This result indicates that if we have at least two annotations for a single debate, then the possibility of having a correct annotation from one of the annotators is pretty high.

\paragraph{Disagreements between the annotators} 
There are generally two types of disagreements between the annotators: (i) same interpretation of the debate but different logic patterns and (ii) different interpretation of the debate.

One example of case (i) is given below:
\begin{itemize} [label={}]
    \item Annotation 1: \emph{\{``homework" is more important than "free time" given the condition/rationale that ``homework can establish basic foundation of studying"\}}
    \item Annotation 2: \emph{\{\{"homework" promote ``establish basic foundation of studying"\} which is more important/severe/has greater weight than \{"homework" suppress "free time"\}\}}.
\end{itemize}
In this example, both annotations have the same interpretation i.e., \emph{homework is more important than free time because it establish basic foundation of studying} but the interpretation is represented differently.

\begin{figure*}[!t]
\begin{center}
\includegraphics[width=16.2cm, height=8.5cm]{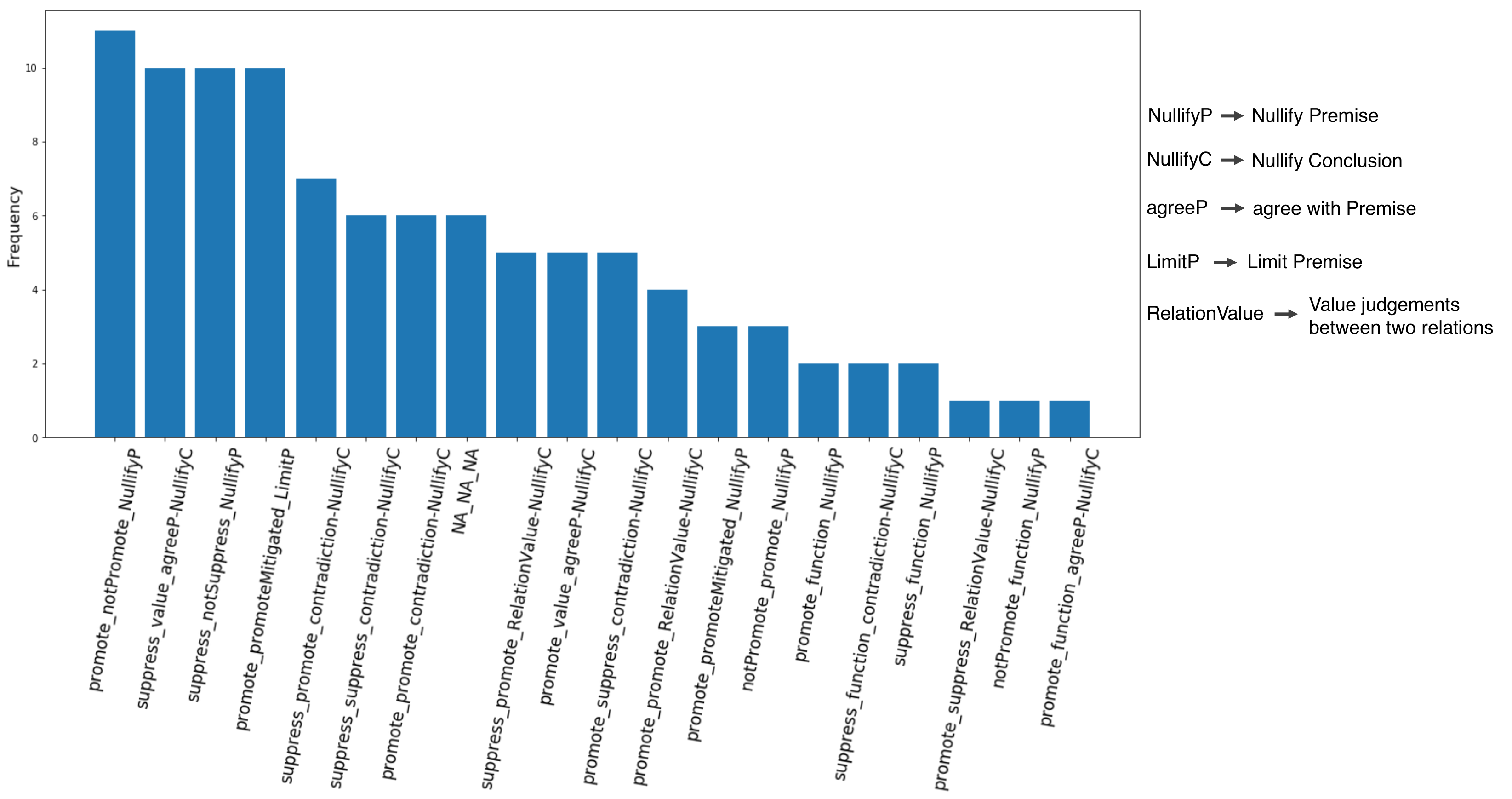} 
\caption{Distribution of logic patterns}
\label{fig:dist-logic}
\end{center}
\end{figure*}

One important factor that we noticed is that in all of the cases of (ii) where annotators have different interpretation of the debates, one of the annotations has been found incorrect. One example of such case is shown in Fig. \ref{Fig:dif-interpret}.
In this example, the interpretation of CA-pattern is different in two annotations i.e., \emph{\{death penalty suppress life imprisonment system which is a bad thing\}} and \emph{\{no death penalty doesn't promote rehabilitation of the criminals\}}. Although both of the annotated patterns are understandable without reading the debates, \emph{Annotation 1} is has been marked as incorrect pattern because in this debate CA doesn't exactly express that life imprisonment is bad, instead it expresses that the reason behind abolishing death penalty is rehabilitation of the criminals but even if we abolish death penalty, it doesn't results in rehabilitation in life imprisonment and \emph{Annotation 1} has failed to capture that notion.


We also see that many disagreement happens regarding 
the choice of text spans. When we manually checked the debates, we have noticed that even when some interpretations are quite the same (case (i)), text spans are different. This is because sometimes two or more sentences express the same meaning and annotators choose text spans from these different sentences. Consider the following example:
\begin{itemize}[label={}]
    \item Annotation 1: \emph{\{``homework" promote ``learns that the way to succeed is by making schedule"\} is more important/severe/has greater weight than \{``homework" suppress "do more of what we really wanted"\}} 
    \item Annotation 2: \emph{\{``homework" promote ``learns the importance of scheduling"\} is more important/severe/has greater weight than \{``homework" suppress ``free time"\}}.
\end{itemize}

In both cases, the text spans \emph{``learns that the way to succeed is by making schedule"} and \emph{``learns the importance of scheduling"} basically have the same interpretation but were chosen from different sentences and therefore are considered as mismatched text spans.

Besides, when CA attacks IA's conclusion, annotators often choose different main points to represent the premise of IA and in such cases text spans don't match.
In the above example we see that the representation of the main points of IA is different i.e., \emph{`do more of what we really wanted'} and \emph{`free time'} since in these annotation CA doesn't attack these premises.

\paragraph{Common Attacking Strategies Captured by the Annotations} 

In order to examine what sort of common rhetorical moves, assumptions or value judgements in attacks are captured by the annotations of two annotators, we look into the distribution of relations and attributes used to annotate the debates. Fig.~\ref{fig:dist-logic} shows this distribution.
From the distribution we observe that the most common logic patterns are `attacking a premise by negating it', `value judgements between two concepts of a premise that leads to agreeing with the premise but denying the conclusion', `providing a way for mitigating the consequence of a premise that leads to agreeing with the premise and nullifying it at the same time' and `providing a contradictory premise that leads to denying the conclusion'.
Moreover, we observe that `value judgement between two causal relations' also happens quite often.

\section{Discussion and Future Work}
In our annotation study, we have observed that although the initial arguments were causal arguments, some logic in the arguments were evaluative judgement e.g., ``death penalty is cruel" or ``A truly just society can do without the death penalty" and the counterarguments focused on those logics. In such cases annotators failed to annotate the attacking strategy. 
In future, we would like to enrich our scheme so that these sort of logics in attack can be captured.
Besides, we plan to have a second annotation for the 145 debates in our corpus that currently have single annotation. 
We also intend to perform a voting between the two annotations so that we can select a single representation for each attacking strategy based on majority voting.
Furthermore, we plan to formulate the task of automatic identification of logic pattern of attacks from given arguments and counterarguments.

We acknowledge the fact that capturing logic pattern of attacks is quite a challenging task, specially when the arguments are long and there is still many room for improvements.

\section{Conclusions}
We proposed LPAttack, a feasible annotation scheme for capturing underlying logic pattern of attacks in arguments.
LPAttack is designed to capture the common strategic moves, assumptions or value judgements during attacks in arguments.
Our annotation study showed that even with a limited set of relations and attributes, we can capture the logic pattern of a wide range of attacks (90\%) in a debate corpus consisting multiple, diverse debate themes.  
The results also showed a moderate inter-annotator agreement (Cohen's $\kappa$=0.49) between two annotators, verifying the feasibility of the proposed scheme.

\section{Bibliographical References}
\label{reference}

\bibliographystyle{lrec2022-bib}
\bibliography{lrecbib}


\section*{Appendix}

\begin{table*}[!t]
\centering
\small

\begin{tabular}{l|p{3.9cm}|p{4.35cm}}
\toprule
\textbf{Relations} & \textbf{Description} & \textbf{Example} \\
\midrule

\raisebox{-.5\height}{\includegraphics[scale=0.07]{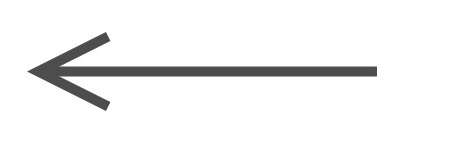}} promote 
& represents something causing/ encouraging another thing 
& no homework \emph{promote} free time \\ [3ex]
\hline & \\[-2.2ex]
 
\raisebox{-.4\height}{\includegraphics[scale=0.07]{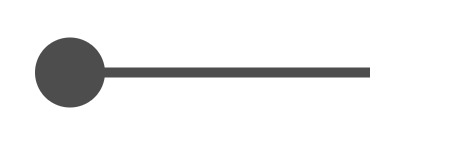}} suppress 
& represents something hindering/ preventing another thing.
& homework suppress free time \\ [3ex]
\hline & \\[-2.2ex]

\raisebox{-.4\height}{\includegraphics[scale=0.07]{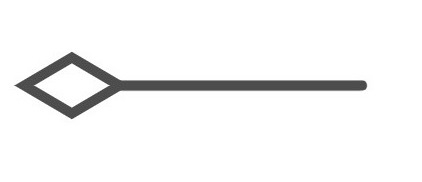}} rationale/condition 
& represents writer's reasoning or justification behind a relation or attribution.
& homework is more important than free time given the \emph{rationale/condition} that homework is part of education \\ [0.5ex]

\hline
\raisebox{-.4\height}{\includegraphics[scale=0.07]{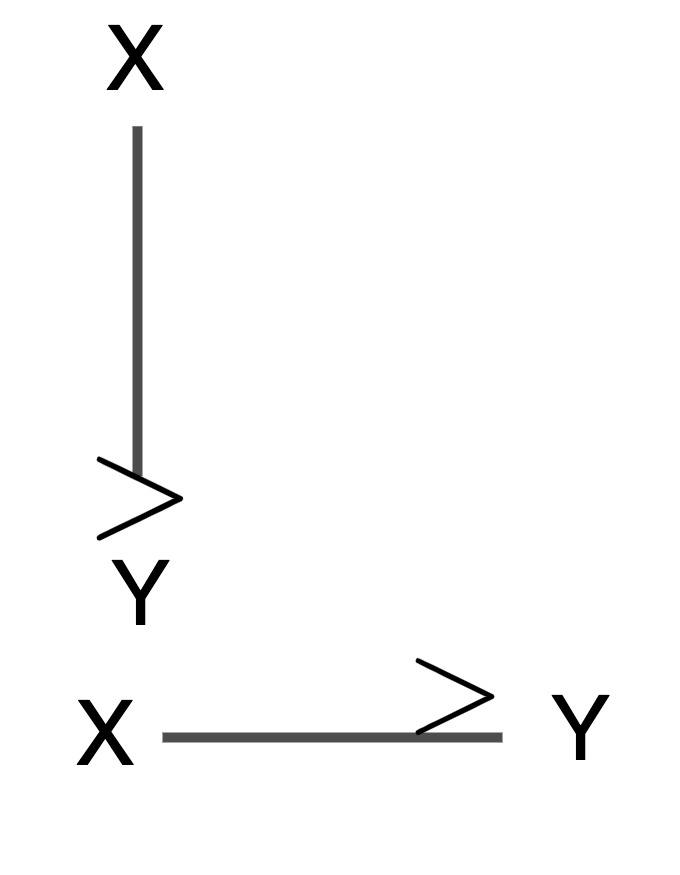}} \makecell{($Y$) is more important/severe/ \\ has greater weight than ($X$) \\ \\ ($X$) is more important/severe/ \\ has greater weight than ($Y$)} 
& \vspace{-1.2cm} (i) represents some relation has higher value than another or (ii) some concept or element has higher value than another 
& \vspace{-1.2cm} (i)\{no homework promote people fail in exam\} which \emph{is more important/ severe/ has greater weight} than \{no homework promote free time\}, (ii) Example in Fig.~\ref{Fig:lpattack-example} \\ [7ex]

\hline& \\[-2.2ex]

\raisebox{-.4\height}{\includegraphics[scale=0.07]{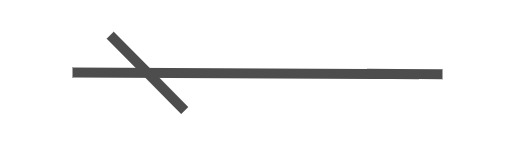}} contradiction 
& represents opposing logics
& \{homework promote ``problems in family”\} \emph{contradicts} \{homework promote good family relation\} \\  [3ex]

\hline& \\[-2.2ex]
\raisebox{-.4\height}{\includegraphics[scale=0.07]{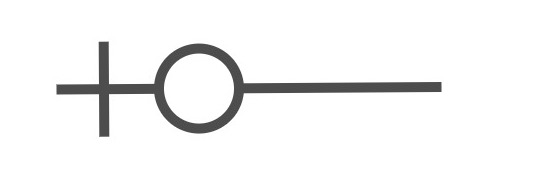}} acknowledgement 
& represents agreement between relations
& Example in Fig.~\ref{Fig:lpattack-example} \\ [3ex]
\hline& \\[-2.2ex]

\raisebox{-.4\height}{\includegraphics[scale=0.07]{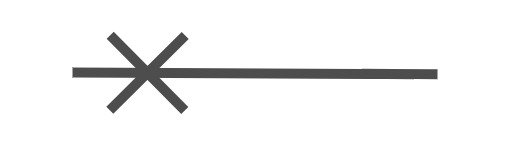}} nullify (attacking relation)
& represents denying a relation or logic 
& Example in Fig.~\ref{Fig:lpattack-example} \\ [3ex]
\hline& \\[-2.2ex]
\raisebox{-.4\height}{\includegraphics[scale=0.07]{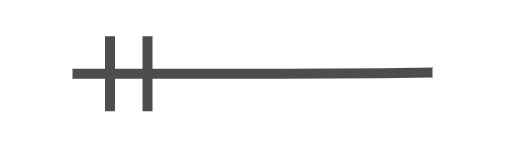}} limit (attacking relation) 
& represents agreeing with and denying a relation at the same time
& \{death penalty promote executioner's suffering can be mitigated given the condition that executioners have a good mental support system\} which \emph{limit} \{death penalty” promote executioner's suffering\}\\ [3ex]
\hline
\rule{0pt}{3ex} \raisebox{-.4\height}{\includegraphics[scale=0.07]{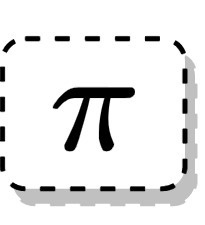}} function 
& represents joining of two or more relations 
& joining the two relations \{homework suppress free time\} and \{free time promote unproductive activities\} would produce the relation \{homework suppress unproductive activities\} \\ [1ex]

\hline
\noalign{\vskip 2mm}
\hline & \\[-1.5ex]
\textbf{Attributes} & \textbf{Description} & \textbf{Example}  \\ [0.5ex]
\hline & \\[-2ex]

\raisebox{-.4\height}{\includegraphics[scale=0.07]{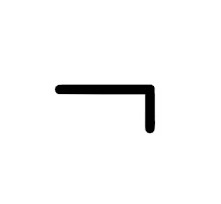}} negation 
& represents negation form of a relation or concept 
& \{homework \emph{doesn’t} promote free time\} or \{\emph{no} homework promote free time\} \\ [3ex]

\hline
\raisebox{-.4\height}{\includegraphics[scale=0.07]{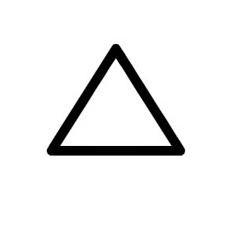}} mitigation 
& represents mitigated form of a relation
& \{death penalty promote executioner's suffering can be \emph{mitigated} given the condition that executioners have a good mental support system\} \\ [5ex]
\hline & \\[-2.2ex]
\hspace{0.1cm} \raisebox{-.4\height}{\includegraphics[scale=0.07]{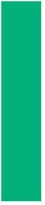}} \space \space \space \space good 
& represents positive feeling of the arguer towards a concept
& \{homework” should be abolished because homework suppress free time\}. Here “free time” is a \emph{good} thing according to the arguer\\ [2ex]

\hline & \\[-2.2ex]
\hspace{0.1cm} \raisebox{-.4\height}{\includegraphics[scale=0.07]{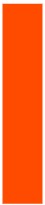}} \space \space \space \space bad 
& represents positive feeling of the arguer towards a concept 
& \{death penalty should be abolished because death penalty promote executioner's suffering \}. Here, ``executioner's suffering” is a \emph{bad} thing according to the arguer \\ [0.5ex] 

\bottomrule

\end{tabular}
\caption{Relations and attributes in LPAttack scheme}
\label{tab:template_set}
\end{table*}

\paragraph{Pre-Study and LPAttack Scheme Design}
The initial design of the annotation schemes, guidelines and the task were developed by the authors of this paper. Then, two non-trained annotators explored the initial designs and helped improving the overall design and the guidelines. 
The primary feedback from the annotators were having a detailed description for each of the relations and attributes, creating categories for them as well as having a prioritization map for these relations and attributes.

During our pre-study, we have noticed is that sometimes the counterargument shows strong opposite sentiment towards an argument e.g., \emph{\{Argument: X is negative; Counterargument: \underline{X is positive}\}} and sometimes the opposite sentiment of the counterargument is not that strong e.g., \emph{\{Argument: X is negative; Counterargument: \underline{X is not negative}\}}. 
However, it usually depends on human perception how ``strong" the opposite sentiment is and might vary from human to human. In order to reduce the complexity and confusion during human annotation, we only keep the representation of the less strong opposite sentiment of counterargument as shown in Fig.~\ref{Fig:base-pattern} of the paper.



\begin{table*}[!t]
\centering
\begin{tabular}{lp{12.5cm}}
\hline
\ & \textbf{Homework should be abolished}\\
\hline
PM-1 & Abolishing homework will give students more free time \\
PM-2 & Forcing students to do homework will make them passive in character \\
PM-3 & It is not good for students to be obliged to study by their teachers or parents \\
PM-4 & Students have memorized the incorrect way to study with homework \\
PM-5 & Schools should take the responsibility for children’s academic skills, not parents at home\\
\hline
\end{tabular}

\vspace{5mm}
\begin{tabular}{lp{12.5cm}}
\hline
\ & \textbf{Death penalty should be abolished}\\
\hline
PM-1 & Death penalty is inhumane punishment \\
PM-2 & Abolishing death penalty will prevent the situation of ending the life of innocent people \\
PM-3 & Because of the high stress on the executioner, death penalty should be abolished \\
PM-4 & The death penalty deprives criminals of the opportunity for rehabilitation \\
PM-5 & Society is being brutalized by the death penalty \\
\hline
\end{tabular}
\caption{Main points of the initial arguments of the debates in the TYPIC corpus \\ for which counterarguments are written}
\label{tab:pm-speech}
\end{table*}

\begin{figure*}[!ht]
\begin{center}
\includegraphics[width=12cm, height=6cm]{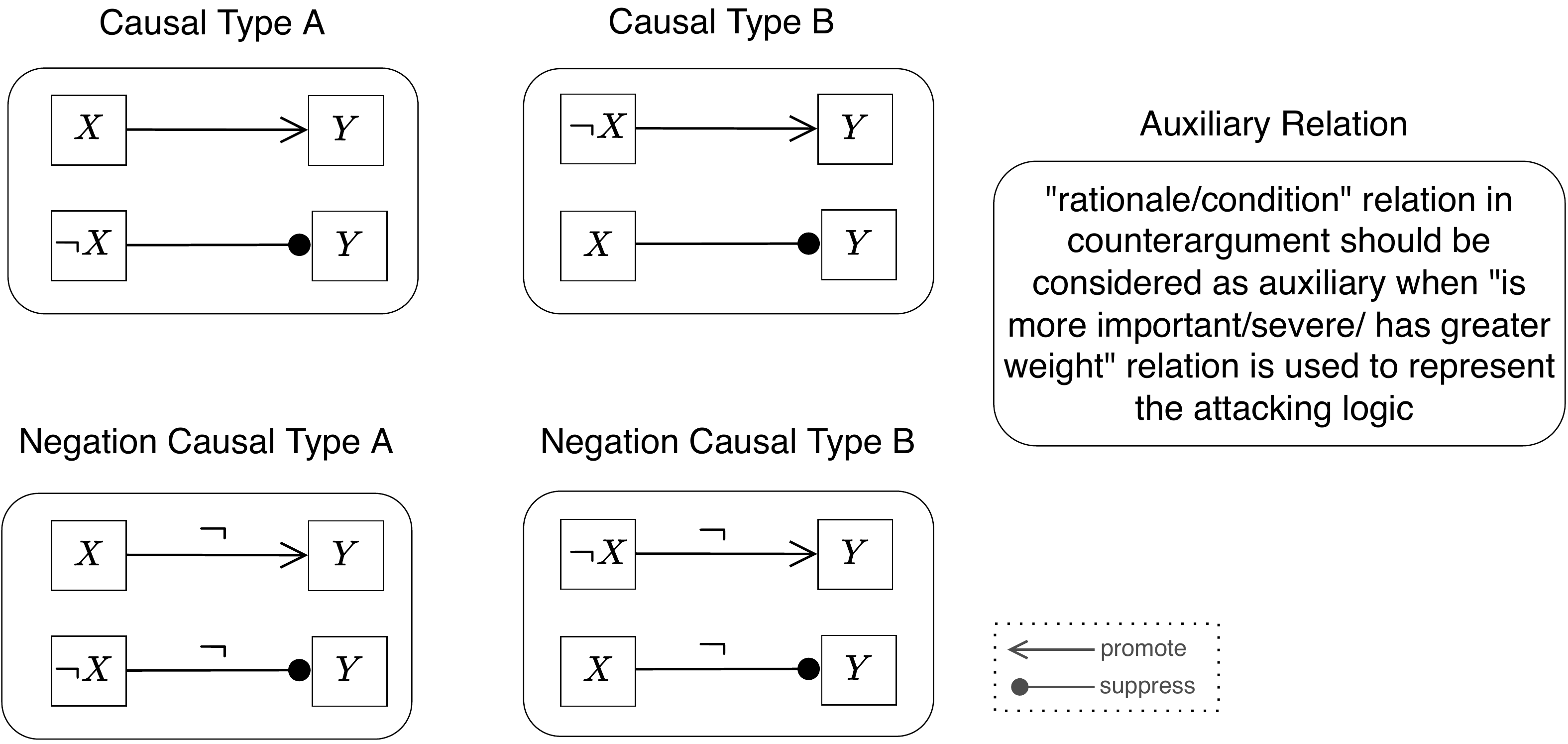} 
\caption{Rules for calculating inter-annotator agreement (IAA)}
\label{Fig:IAA_rules}
\end{center}
\end{figure*}

\paragraph{Source Data}
For our pre-study and annotation study, we utilize the debates from TYPIC dataset~\cite{naito2022}. This dataset has 1,000 parliamentary style debates where given a topic, two opposing teams i.e., Prime Minister (PM) and the Leader of the Opposition (LO) argue by taking a position in favor and against the topic respectively. 
In each debate, the PM speech acts as the \emph{initial argument} and the LO speech acts as the \emph{counterargument}.
The corpus consists 10 PM speeches which are written on two topics: \emph{``Homework should be abolished"} and \emph{``Death penalty should be abolished"}.
Table \ref{tab:pm-speech} shows the main points of these PM speeches. For each of the PM speech, there are 100 LO speeches.
The arguments of 8 PM speeches out of 10 are causal arguments (underlined in the table).

Since our scheme is designed around two causal relations ``promote" and ``suppress", when we choose debates, we only choose these 8 PM speeches (initial arguments) for annotations whose arguments are causal arguments and then we randomly choose LO speeches (counterarguments) associated with these PM speeches.

\paragraph{Rules for calculating IAA:}
One factor to consider during the IAA calculation is that in our scheme, we kept the flexibility of human representation i.e., the same interpretation can be represented in a slightly different way e.g., ``no homework promote free time" has the same meaning as ``homework suppress free time" but they are different representations.
In order to handle such different representations that generally have the same meaning, we create some rules to consider these different representation as same.
Fig.~\ref{Fig:IAA_rules} show these rules.
As shown in the figure, we consider the representations ``no X promote Y" and ``X suppress Y" as same (marked as \emph{Causal Type A}) since they have the same meaning.
One of our rules consider rationale/condition relation as auxiliary in certain cases where having or not having it doesn't affect the understanding of the logic much. For example, in the case of the logic \{\{no homework promote people fail in exam \emph{given the rationale/condition that not doing homework will lead to lack of preparation\}} which is more important or severe or has greater weight than \{no homework promote free time\}\}, even if we remove the rationale/condition relation, the interpretation is understandable. We ignore this relation in such cases during the calculation of IAA.


\begin{figure*}[b]
\begin{center}
\includegraphics[width=13.5cm, height=15cm]{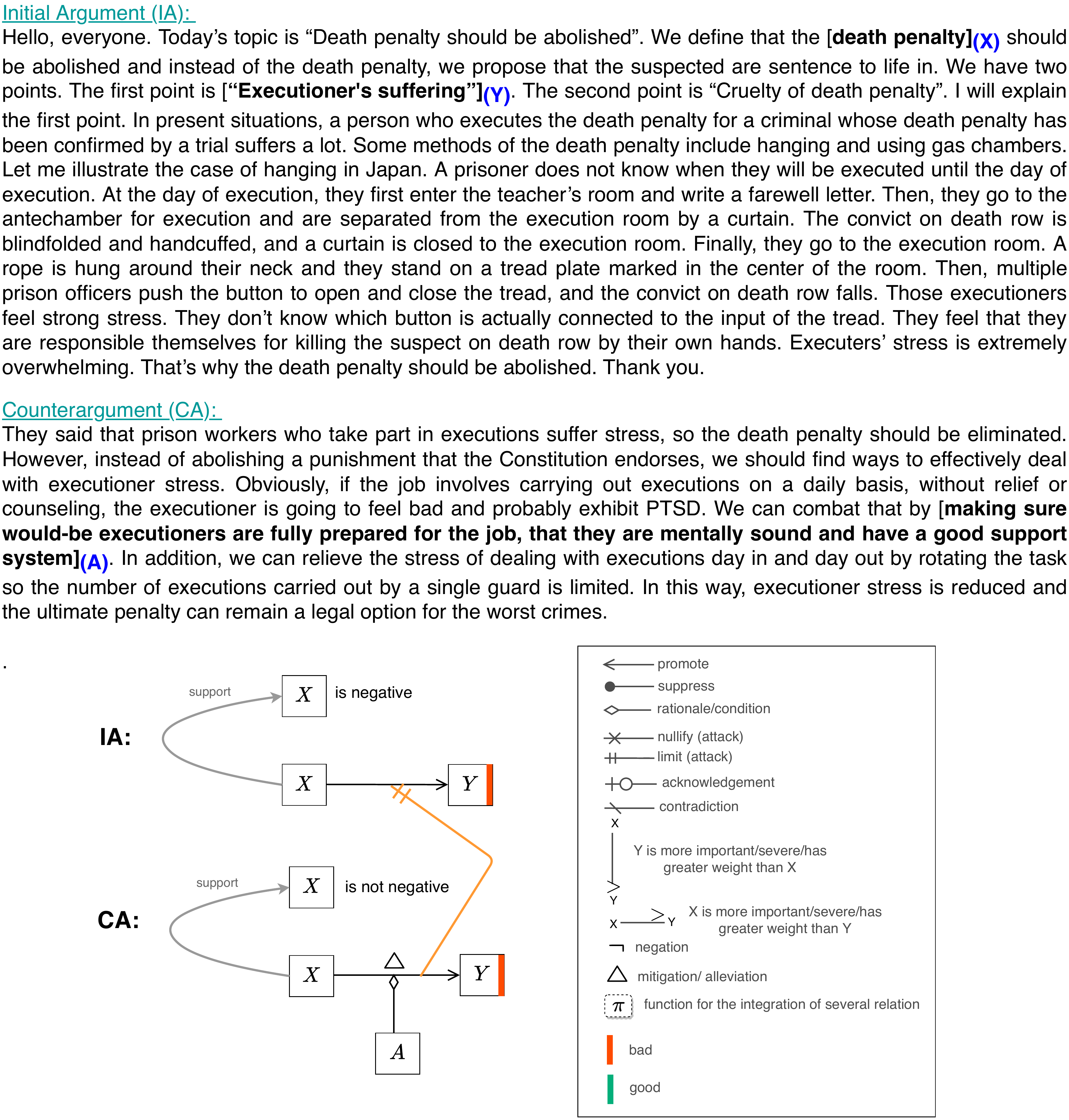} 
\label{fig:dif-interpret}
\end{center}
\caption{Annotation example of logic pattern of attack of a debate}
\label{ann-ex3}
\end{figure*}

\begin{figure*}[!t]
\begin{center}
\includegraphics[width=13cm, height=10.8cm]{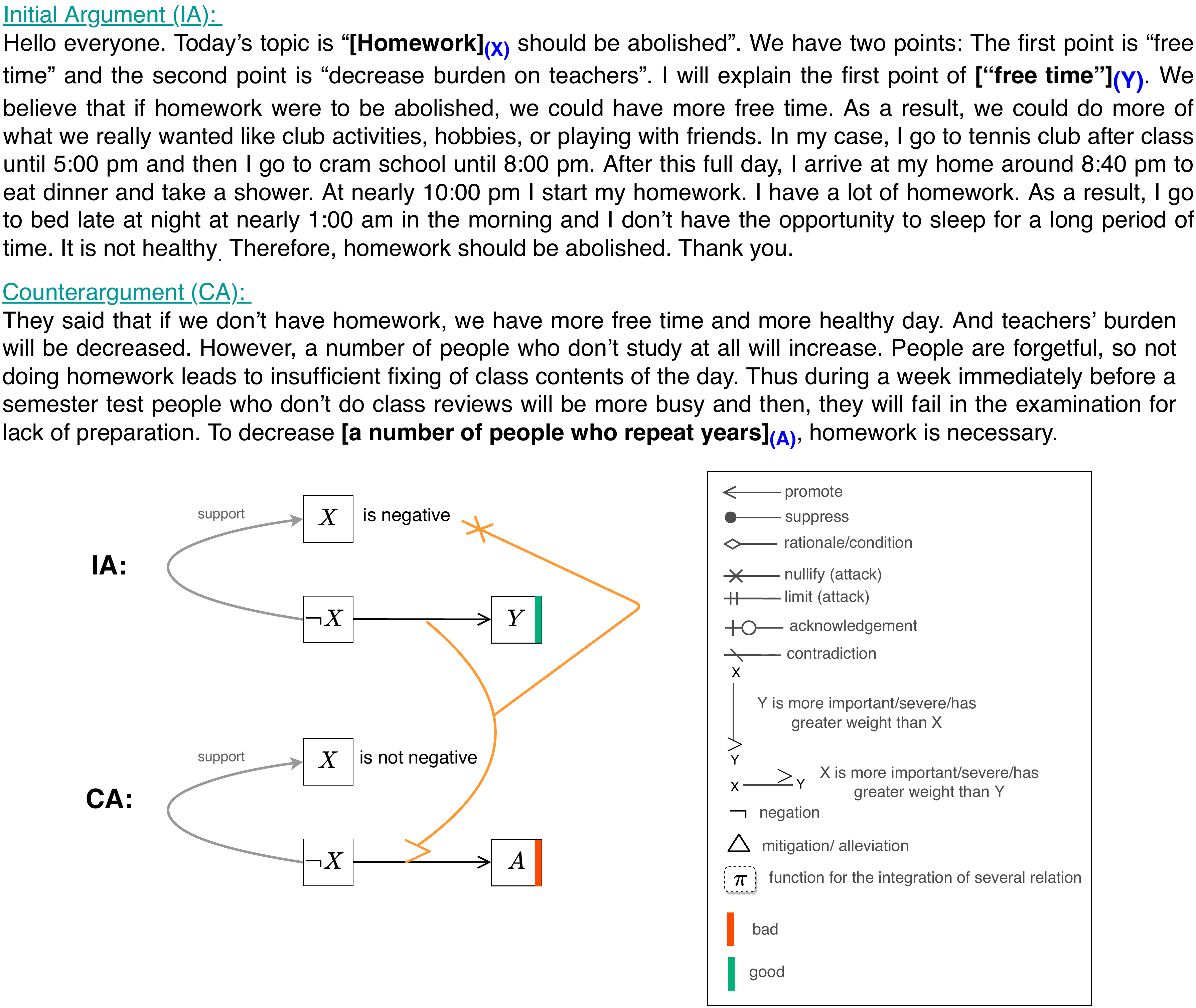} 
\label{fig:dif-interpret}
\end{center}
\caption{Annotation example of logic pattern of attack of a debate}
\label{ann-ex2}
\end{figure*}

\begin{figure*}[!h]
\begin{center}
\includegraphics[width=12.8cm, height=11.3cm]{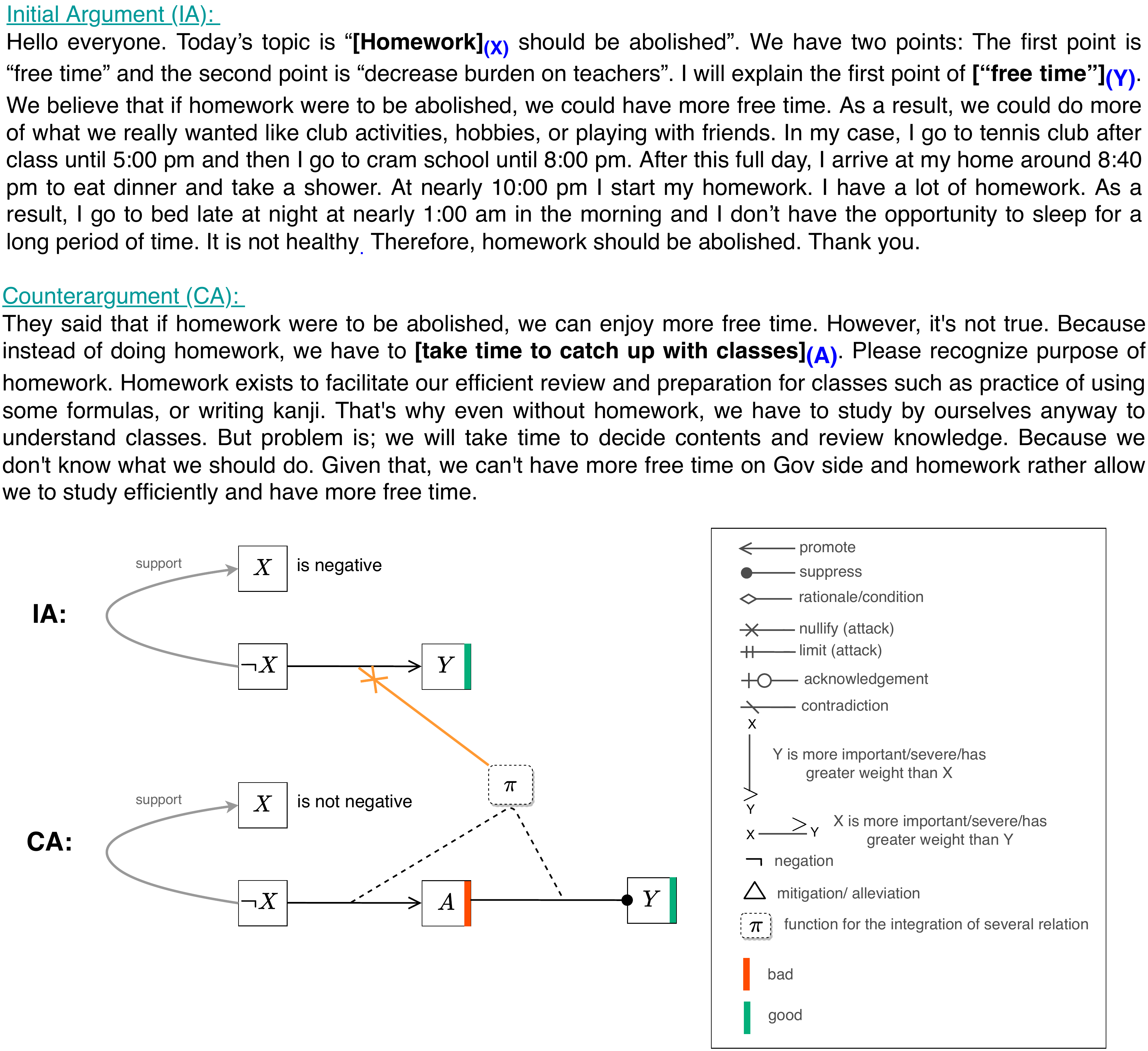} 
\label{fig:dif-interpret}
\end{center}
\caption{Annotation example of logic pattern of attack of a debate}
\label{ann-ex1}
\end{figure*}

\paragraph{Annotation examples of logic pattern of attacks}
In order to provide a better understanding of what the annotated logic patterns look like and what sort of text spans are chosen from the given arguments, we provide some annotation examples of the logic pattern of attacks in Fig. \ref{ann-ex3}, \ref{ann-ex2} and \ref{ann-ex1}.

\end{document}